%% file: main.tex
\documentclass[runningheads]{llncs}

\usepackage[T1]{fontenc}

\usepackage{graphicx}
\graphicspath{{figures/}}
\usepackage{caption}
\usepackage{booktabs}
\usepackage{array}
\usepackage{tabularx}
\newcolumntype{L}{>{\raggedright\arraybackslash}X}
\usepackage{siunitx}
\usepackage{amsmath}
\usepackage{amssymb}
\usepackage{mathtools}    

\usepackage[expansion=false]{microtype}
\usepackage{xcolor}

\usepackage[hidelinks]{hyperref}

\let\savedthebibliography\thebibliography
\renewcommand{\thebibliography}[1]{%
  \savedthebibliography{#1}%
  \scriptsize
  \setlength{\itemsep}{0pt plus 0.5pt}%
  \setlength{\parsep}{0pt}%
}

\begin{document}

\title{SketchMamba: A Lightweight State-Space Model for Joint Progressive Sketch Classification and Stroke Auto-Completion}
\titlerunning{SketchMamba}

\author{%
Kavish Jhaveri\inst{1}\orcidID{0009-0002-2951-3232} \and
Arya Shah\inst{2}\orcidID{0000-0002-2649-1835}%
}
\authorrunning{K. Jhaveri et al.}

\institute{%
The National High School, Ahmedabad, India\\
\email{kavishparthivjhaveri@gmail.com}
\and
Indian Institute of Technology, Gandhinagar, India\\
\email{arya.shah@iitgn.ac.in}%
}

\maketitle

\input{sections/00_abstract}
\input{sections/01_introduction}
\input{sections/02_related_work}

\input{sections/03_methodology}

\input{sections/04_experiments}

\input{sections/05_results}
\input{sections/06_conclusion}
\input{sections/07_credits}

\def\doi#1{doi:\detokenize{#1}}
\bibliographystyle{splncs04}
\bibliography{references}

\end{document}

%% file: sections/00_abstract.tex
\begin{abstract}
Existing vector-sketch models treat recognition and generation as separate
tasks, leaving a gap for streaming interfaces that must understand a
drawing as it is being made. We present SketchMamba, a single causal
sequence model that continuously classifies a sketch from any partial
prefix while simultaneously generating its continuation. We achieve this
by applying a dense per-step classification loss to a selective state-space
backbone. Evaluated on a 58-class subset of the Quick, Draw! dataset,
SketchMamba yields 94.93\% final-step accuracy and a progressive-accuracy
Area Under the Curve (AUC) of 0.706, crossing 90\% of its final accuracy
by the time 70\% of the strokes are drawn. In a matched-budget comparison,
the 1.55 million-parameter backbone ties a causal Transformer while
outperforming recurrent and convolutional baselines. Ablations confirm
that the dense supervision regime, rather than the architecture alone,
drives the early-prediction capability. The results demonstrate that a
single causal hidden state can unify progressive recognition and
autoregressive generation without auxiliary encoders or task-specific
branching.

\keywords{Vector Sketch \and State Space Models \and Progressive Recognition \and Sequence Modelling}
\end{abstract}

%% file: sections/01_introduction.tex
\section{Introduction}
\label{sec:intro}

Digital ink interfaces on mobile devices make recording stroke sequences
routine. Large-scale datasets like Quick, Draw!~\cite{googlequickdraw}
permit training robust sequence models for these data. Treating a drawing
as an autoregressive sequence of pen displacements was established for
online handwriting by Graves~\cite{graves2013seq} and adapted for sketches
by Sketch-RNN~\cite{ha2018sketchrnn}, using a recurrent decoder with a
Gaussian-mixture output head to synthesise continuations. As streaming
sketch interfaces become standard, models must process and describe these
drawings as they are actively being drawn.

Existing work treats recognition and generation separately. Recognition
models like Sketchformer~\cite{sketchformer} and
Sketch-BERT~\cite{sketchbert} use Transformer backbones to classify
finished sketches, leaving per-prefix recognition undefined. Generation
models like Sketch-RNN~\cite{ha2018sketchrnn},
SketchHealer~\cite{sketchhealer}, and SketchKnitter~\cite{sketchknitter}
treat class identity as a conditioning input or a downstream task. No
prior vector-sketch model supervises a single causal hidden state at every
step to simultaneously classify the sketch from any partial prefix and
autoregressively predict the next stroke.

\begin{figure}[t]
  \centering
  \includegraphics[width=0.58\linewidth]{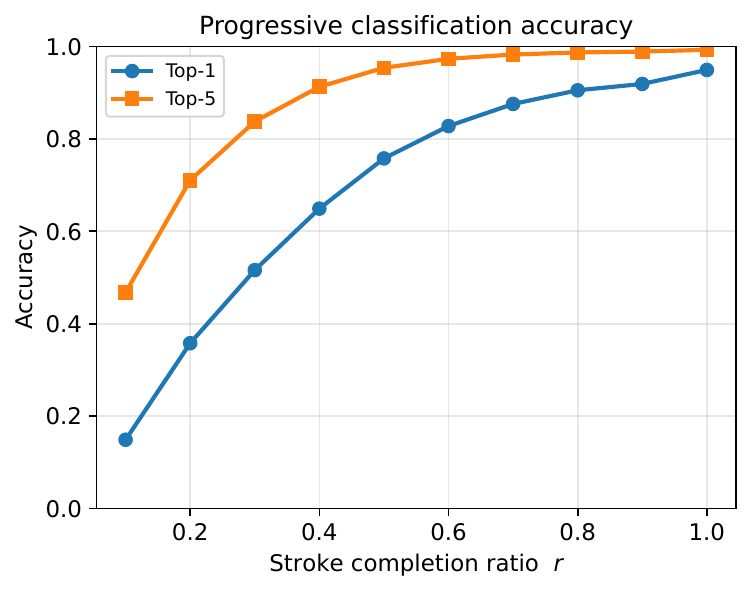}
  \caption{Progressive classification on the 145{,}000-sketch
  Quick, Draw! test split (58 classes). SketchMamba reaches 75.8\%
  Top-1 by half a sketch and over 91\% Top-5 by 40\% of the strokes;
  the area under this curve, normalised to the unit interval, is
  0.706. The curve is read at any prefix length~$r$ from the same
  causal hidden state used to generate stroke continuations.}
  \label{fig:teaser}
\end{figure}

\emph{SketchMamba} addresses this using a causal selective state-space
backbone~\cite{mamba2023,gu2022s4} that drives three per-step heads: a
58-class classifier, a 20-component bivariate Gaussian-mixture stroke
head~\cite{bishop1994mdn}, and a three-way pen-state head. Trained jointly
on Quick, Draw!, the model uses dense per-step supervision, applying
classification cross-entropy at every time step rather than only at the
final token. This supervision forces a single causal state to operate as
both an early-prediction recogniser and a conditioning signal for
autoregressive generation.

On the 145{,}000-sketch test split, SketchMamba reaches 94.93\% Top-1
final-step accuracy. Its progressive-accuracy curve yields a normalised
area of 0.706, crossing 90\% of final accuracy by the time 70\% of strokes
are drawn (Fig.~\ref{fig:teaser}). At a matched 1.5 to 1.9 million parameter
budget, the Mamba backbone matches a Transformer~\cite{vaswani2017} on
Top-1 accuracy and outperforms LSTM, GRU, and dilated CNN baselines.
Replacing the dense per-step loss with a final-step-only loss collapses
the progressive curve area from 0.706 to 0.074, confirming the supervision
regime drives early prediction. The architecture is lightweight, running
in 1.5~ms per sequence on an NVIDIA L40S GPU.

Dense per-step supervision on a causal state-space backbone produces an
early-stroke recogniser and a generator from the same hidden state. To
avoid confounding architectural and training-recipe
variation~\cite{bouthillier2021variance}, the backbone study and ablation
matrix operate within a strictly fixed training pipeline.

%% file: sections/02_related_work.tex
\section{Literature Review}
\label{sec:related}

\begin{table}[t]
  \centering
  \caption{Comparison with representative prior work on vector-sketch
  modelling. ``Joint state'' indicates whether a single causal hidden
  state is shared between recognition and stroke generation;
  ``Per-step'' indicates whether the recognition loss is applied at
  every step. ``Stream'' means the model can predict class at any
  prefix without recomputing from scratch. (cond.\ = class is a
  conditioning input, impl.\ = implicit via RL, (\checkmark) = partial).}
  \label{tab:relwork-compare}
  \small
  \setlength{\tabcolsep}{3pt}
  \resizebox{\linewidth}{!}{
  \begin{tabular}{@{}ll l c c c c c@{}}
    \toprule
    Work & Modality & Backbone & Recog. & Gen. & Joint state & Per-step & Stream \\
    \midrule
    Sketch-a-Net~\cite{yu2017sketchanet}  & raster & CNN            & \checkmark & $\times$   & $\times$ & $\times$ & $\times$ \\
    Sketch-RNN~\cite{ha2018sketchrnn}     & vector & LSTM           & cond.      & \checkmark & $\times$ & $\times$ & \checkmark \\
    Sketch-R2CNN~\cite{li2018sketchr2cnn} & hybrid & RNN+CNN        & \checkmark & $\times$   & $\times$ & $\times$ & (\checkmark) \\
    Sketchformer~\cite{sketchformer}      & vector & Transformer    & \checkmark & (\checkmark) & $\times$ & $\times$ & $\times$ \\
    Sketch-BERT~\cite{sketchbert}         & vector & bidir.~Trans.  & \checkmark & (\checkmark) & $\times$ & $\times$ & $\times$ \\
    BezierSketch~\cite{das2020beziersketch}& vector & RNN+B\'ezier   & $\times$   & \checkmark & $\times$ & $\times$ & \checkmark \\
    CoSE~\cite{aksan2020cose}             & vector & hier.~Trans.   & $\times$   & \checkmark & $\times$ & $\times$ & (\checkmark) \\
    DeepSVG~\cite{carlier2020deepsvg}     & SVG    & hier.~Trans.   & $\times$   & \checkmark & $\times$ & $\times$ & $\times$ \\
    SketchHealer~\cite{sketchhealer}      & graph  & GNN+LSTM       & $\times$   & \checkmark & $\times$ & $\times$ & \checkmark \\
    SketchKnitter~\cite{sketchknitter}    & vector & diffusion      & $\times$   & \checkmark & $\times$ & $\times$ & $\times$ \\
    Pixelor~\cite{bhunia2020pixelor}      & vector & RNN+RL         & impl.      & \checkmark & $\times$ & $\times$ & \checkmark \\
    \midrule
    \textbf{SketchMamba (ours)}           & \textbf{vector} & \textbf{causal Mamba} & \textbf{\checkmark} & \textbf{\checkmark} & \textbf{\checkmark} & \textbf{\checkmark} & \textbf{\checkmark} \\
    \bottomrule
  \end{tabular}
  }
\end{table}

We organise the related work along the four axes that the present paper
touches: vector-sketch understanding and generation, mainstream sequence
backbones, the recent state-space-model lineage, and the methodology
context that frames our matched-budget evaluation. Within each axis we
synthesise rather than enumerate; Table~\ref{tab:relwork-compare}
summarises how the most directly comparable prior systems differ from
SketchMamba on the dimensions that matter for streaming sketch
interfaces.

\subsection{Vector sketch understanding and generation}
\label{ssec:rel-sketch}

Two largely disjoint threads have shaped vector-sketch modelling. The
recognition thread treats the sketch as a static object. Early work
rasterised strokes for CNNs~\cite{yu2017sketchanet}; subsequent systems
combined stroke-level RNNs with CNNs~\cite{li2018sketchr2cnn}, exploited
graph structure~\cite{yang2021sketchgnn}, or scaled via deep
hashing~\cite{xu2018sketchmate} and fine-grained
retrieval~\cite{sangkloy2016sketchy}. Transformer encoders enabled
Sketchformer~\cite{sketchformer} and the bidirectional, self-supervised
Sketch-BERT~\cite{sketchbert}. These produce a single classification
score on the finished sketch, leaving the per-prefix recognition curve
undefined.

The generation thread treats the sketch as a stochastic sequence.
Sketch-RNN~\cite{ha2018sketchrnn} models stroke offsets with a bivariate
Gaussian-mixture head~\cite{bishop1994mdn}, building on Graves's LSTM
handwriting decoder~\cite{graves2013seq}. Subsequent generators explore
CNN-RNN encoders~\cite{chen2017sketchpix2seq}, B\'ezier
embeddings~\cite{das2020beziersketch}, compositional
strokes~\cite{aksan2020cose}, hierarchical Transformers for
SVGs~\cite{carlier2020deepsvg}, lattice
grids~\cite{qi2021sketchlattice}, neural-ODEs~\cite{das2022sketchode},
abstract priors~\cite{yang2021sketchaa}, creative-sketch
GANs~\cite{ge2020doodlergan,bhunia2022doodleformer}, graph-to-sequence
healing~\cite{sketchhealer,su2022generative}, diffusion
samplers~\cite{sketchknitter}, and CLIP-guided
rasterisers~\cite{vinker2022clipasso}. Class identity is treated as a
one-hot conditioning input rather than a prediction target.

A few works sit closer to our streaming objective.
Pixelor~\cite{bhunia2020pixelor} trains an RL agent to race a human to
recognisability but does not formalise dense per-step supervision;
Sketch-R2CNN~\cite{li2018sketchr2cnn} attends over points before
rasterising, breaking causal sequence generation. The gap remains a
single causal backbone supervised at every step for both tasks.

\subsection{Sequence backbones}
\label{ssec:rel-backbones}

Four backbone families dominate causal sequence modelling. Recurrent
networks (LSTM~\cite{hochreiter1997lstm}, GRU~\cite{cho2014gru}) have
$O(L)$ inference cost but suffer optimisation difficulties on long
horizons. Causal dilated 1-D convolutions (WaveNet~\cite{oord2016wavenet},
TCN~\cite{bai2018tcn}) replace recurrence with fixed receptive fields.
Transformers~\cite{vaswani2017} use global self-attention but scale
quadratically, motivating segment-recurrent~\cite{dai2019transformerxl},
sparse~\cite{beltagy2020longformer}, and
linear~\cite{katharopoulos2020linear} variants. Our matched-budget
comparison is justified by benchmarking work showing that training
recipes and scaling rules, rather than architecture, often dominate
cross-paper deltas~\cite{bello2021revisiting,liu2022convnext}, with
optimisers as a confound~\cite{sivaprasad2020optimizer,bouthillier2021variance}.

\subsection{State-space models for sequences}
\label{ssec:rel-ssm}

Structured state-space models reframe long-sequence modelling as a
learnable linear recurrence computed by a parallel scan. Originating with
HiPPO~\cite{gu2020hippo} and S4~\cite{gu2022s4}, subsequent variants
simplify the parameterisation~\cite{smith2023s5}, close the language
gap~\cite{fu2023h3}, or replace attention with long
convolutions~\cite{poli2023hyena}. Mamba~\cite{mamba2023} introduces
input-dependent selectivity with a hardware-aware scan;
Mamba-2~\cite{dao2024mamba2} formalises a duality with attention. The
primitive is conceptually close to other linear-attention
recurrences~\cite{peng2023rwkv,sun2023retnet} and provides exactly what
we need: a causal hidden state at every step.

\subsection{Training methodology and evaluation context}
\label{ssec:rel-method}

Our joint objective is an instance of multi-task hard parameter
sharing~\cite{caruana1997multitask,ruder2017multitask}. The optimisation
recipe (AdamW, cosine warmup, RMSNorm, EMA, bf16) follows standard
practice~\cite{loshchilov2019adamw,loshchilov2017sgdr,zhang2019rmsnorm,izmailov2018swa}
and numerical stability findings~\cite{kalamkar2019bf16}.
The progressive-accuracy curve operationalises early classification for
time series~\cite{xing2009early}, and macro-F1 is used for class
imbalance~\cite{sokolova2009measures}.

%% file: sections/03_methodology.tex
\section{Methodology}
\label{sec:methodology}

SketchMamba is a single causal sequence model that jointly performs sketch
recognition and autoregressive sketch generation. It reads drawing
actions one at a time and maintains a single hidden state, from which it
continuously predicts the overall class of the sketch while simultaneously
predicting the spatial coordinates and pen state of the next stroke.

\subsection{Data Representation}
\label{ssec:meth-data}

We adopt the canonical five-element vector representation introduced by
Sketch-RNN~\cite{ha2018sketchrnn}. A sketch is a sequence of $T$ points,
where each point is defined as:
\begin{equation}
  \mathbf{x}_t = (\Delta x_t, \Delta y_t, p_{1,t}, p_{2,t}, p_{3,t})
\end{equation}
The first two elements, $\Delta x_t$ and $\Delta y_t$, encode the spatial
offset from the previous point. The remaining three elements form a one-hot
vector representing the state of the pen: $p_{1,t}$ indicates the pen is
touching the paper and drawing a stroke, $p_{2,t}$ indicates the pen is
lifted to move to a new position, and $p_{3,t}$ marks the end of the
drawing. 

\subsection{Model Architecture}
\label{ssec:meth-arch}

The architecture consists of a linear input embedding, a stack of causal
selective state-space blocks, and three parallel output heads that share
the resulting hidden state.

\subsubsection{Backbone}
At each step $t$, a linear layer projects the five-dimensional input
$\mathbf{x}_t$ into a $D$-dimensional continuous embedding. This sequence
is processed by a Mamba~\cite{mamba2023} backbone. As a selective SSM,
Mamba maps the input sequence to a hidden state sequence $\mathbf{h}_t$
through a linear ODE whose discretised parameters are data-dependent. This
selectivity allows the model to compress the drawing history adaptively,
filtering out noisy intermediate movements while retaining structural invariants.
Because the recurrence is strictly causal, $\mathbf{h}_t$ depends only on
$\mathbf{x}_{1 \dots t}$.

\subsubsection{Task Heads}
The shared representation $\mathbf{h}_t$ is passed to three lightweight
parallel heads:
\begin{enumerate}
  \item \textbf{Classification Head:} A linear projection mapping $\mathbf{h}_t$
  to a 58-dimensional logit vector $\hat{y}_t$, representing the predicted
  distribution over the Quick, Draw! categories given the prefix up to step $t$.
  \item \textbf{Stroke Spatial Head:} Following standard practice for continuous
  coordinate generation~\cite{bishop1994mdn,ha2018sketchrnn}, this head outputs
  the parameters of a bivariate Gaussian mixture model (GMM) with $K=20$
  components. A linear layer maps $\mathbf{h}_t$ to a $6K$-dimensional vector
  controlling the mixture weights $\Pi$, means $(\mu_x, \mu_y)$, standard
  deviations $(\sigma_x, \sigma_y)$, and correlation $\rho$ for the next
  spatial offset $(\Delta x_{t+1}, \Delta y_{t+1})$.
  \item \textbf{Pen State Head:} A linear projection mapping $\mathbf{h}_t$
  to three categorical logits for the next pen state $(p_{1,t+1}, p_{2,t+1}, p_{3,t+1})$.
\end{enumerate}

\subsection{Dense Per-Step Supervision}
\label{ssec:meth-loss}

The central methodological contribution of SketchMamba is the dense
application of the classification objective. In standard sequence
classification, cross-entropy is computed only at the final step $T$. We
instead supervise the classification head at every valid time step. 
The total loss is a weighted sum of the recognition and generation objectives:
\begin{equation}
  \mathcal{L} = \lambda_{\text{cls}} \mathcal{L}_{\text{cls}} + 
                \lambda_{\text{stroke}} \mathcal{L}_{\text{stroke}} + 
                \lambda_{\text{pen}} \mathcal{L}_{\text{pen}}
\end{equation}

The classification loss $\mathcal{L}_{\text{cls}}$ averages the cross-entropy
between the ground-truth sketch category $y^*$ and the per-step predictions
$\hat{y}_t$ over the sequence length:
\begin{equation}
  \mathcal{L}_{\text{cls}} = \frac{1}{T} \sum_{t=1}^{T} \text{CrossEntropy}(\hat{y}_t, y^*)
\end{equation}
By forcing the shared hidden state $\mathbf{h}_t$ to predict the final
category from early prefixes, this objective naturally structures the state
space for progressive recognition.

The generative objectives are applied in a standard teacher-forced manner.
The stroke loss $\mathcal{L}_{\text{stroke}}$ is the negative log-likelihood
of the true next offset $(\Delta x_{t+1}, \Delta y_{t+1})$ under the predicted
bivariate GMM. The pen loss $\mathcal{L}_{\text{pen}}$ is the cross-entropy
between the predicted pen-state logits and the true next pen state
$(p_{1,t+1}, p_{2,t+1}, p_{3,t+1})$.

\subsection{Streaming Inference}
\label{ssec:meth-inference}

At inference time, SketchMamba operates as a true streaming system. When a
user draws a stroke, the new point $\mathbf{x}_t$ is ingested in $O(1)$ time
to update the state-space recurrence. The classification head can be queried
instantly to retrieve $\hat{y}_t$, providing a live recognition readout. If
generation is required, the spatial and pen heads are sampled to produce
$\hat{\mathbf{x}}_{t+1}$, which is fed back into the model autoregressively
without recalculating the prefix. This dual capacity is enabled entirely by
the causal architecture and the dense per-step training regime.

%% file: sections/04_experiments.tex
\section{Experiments}
\label{sec:experiments}

We design our experiments to isolate the impact of the causal backbone
and the dense per-step supervision regime. To avoid confounding variables,
all models are trained and evaluated within a single unified pipeline.

\subsection{Dataset and Preprocessing}
\label{ssec:exp-data}

We use the canonical 58-class Quick, Draw! subset~\cite{googlequickdraw,ha2018sketchrnn}.
The data is split into 4.06~million training, 145{,}000 validation, and
145{,}000 test sketches. We apply Ramer-Douglas-Peucker (RDP) simplification
with an epsilon of 2.0 and scale spatial offsets to zero mean and unit
variance. Sequences are capped at a maximum length of $T=200$ points.

\subsection{Matched-Budget Baselines}
\label{ssec:exp-baselines}

To evaluate the state-space backbone, we compare SketchMamba against four
alternative causal sequence operators at a matched parameter budget of
roughly 1.5 to 1.9~million parameters. The baselines substitute the Mamba
blocks for:
\begin{itemize}
    \item \textbf{LSTM~\cite{hochreiter1997lstm} \& GRU~\cite{cho2014gru}:} Standard recurrent networks ($D=512$, 2 layers).
    \item \textbf{CNN1D~\cite{oord2016wavenet,bai2018tcn}:} Causal dilated convolutions with exponentially increasing dilation factors ($D=256$, 8 layers).
    \item \textbf{Transformer~\cite{vaswani2017}:} A standard causal self-attention decoder ($D=256$, 4 layers, 8 heads).
\end{itemize}
All models share the exact same embedding layer, task heads, and training
hardware.

\subsection{Training Protocol and Metrics}
\label{ssec:exp-protocol}

Table~\ref{tab:hyperparams} summarises the fixed training configuration.
We evaluate recognition performance using final-step Top-1 and Top-5
accuracy, alongside the Area Under the Curve (AUC) of the progressive
accuracy measured at 10\% increments of the drawing length. Generation
fidelity is measured via the negative log-likelihood (NLL) of the stroke
spatial coordinates and pen state predictions.

\begin{table}[t]
  \centering
  \caption{Experimental setup and hyperparameter configuration. These
  settings are held constant across all backbone baselines.}
  \label{tab:hyperparams}
  \small
  \setlength{\tabcolsep}{4pt}
  \renewcommand{\arraystretch}{0.95}
  \begin{tabular}{@{}ll@{}}
    \toprule
    \textbf{Hyperparameter} & \textbf{Value} \\
    \midrule
    Dataset & Quick, Draw! (58 classes) \\
    Training samples & 4.06M \\
    Max sequence length & $T=200$ \\
    Optimizer & AdamW ($\beta_1=0.9, \beta_2=0.95$) \\
    Peak learning rate & $5 \times 10^{-4}$ \\
    LR schedule & Cosine with 5000 warmup steps \\
    Batch size & 1024 \\
    Weight decay & 0.1 \\
    Epochs & 50 \\
    Loss weights $(\lambda_{\text{cls}}, \lambda_{\text{stroke}}, \lambda_{\text{pen}})$ & (1.0, 1.0, 0.5) \\
    EMA decay & 0.9999 \\
    Precision & bf16 mixed \\
    \bottomrule
  \end{tabular}
\end{table}

%% file: sections/05_results.tex
\section{Results and Discussion}
\label{sec:results}

This section reports recognition accuracy, generative fidelity, and
inference latency, followed by an ablation of the supervision regime and an
analysis of model limitations.

\subsection{Matched-Budget Recognition}
\label{ssec:res-backbones}

SketchMamba achieves 94.93\% final-step Top-1 accuracy on the Quick, Draw!
test split. In the matched-budget evaluation (Table~\ref{tab:results-backbones}),
the Mamba backbone ties a causal Transformer while using 17\% fewer parameters.
Both models outperform the LSTM, GRU, and CNN1D baselines by margins of 1.5
to 6.6 accuracy points. Under dense per-step supervision, SketchMamba yields
a progressive-accuracy Area Under the Curve (AUC) of 0.706, crossing 90\% of
its final accuracy by the time 70\% of the sketch is drawn.

\begin{table}[t]
  \centering
  \caption{Matched-budget recognition performance on the 58-class test split.
  The SSM matches the causal Transformer while remaining highly parameter-efficient.}
  \label{tab:results-backbones}
  \small
  \setlength{\tabcolsep}{4pt}
  \begin{tabular}{@{}lccc@{}}
    \toprule
    \textbf{Model} & \textbf{Params} & \textbf{Top-1 Acc.} & \textbf{Prog. AUC} \\
    \midrule
    LSTM / GRU / CNN1D & 1.5M -- 1.9M & 88.33\% -- 93.43\% & -- \\
    Causal Transformer & 1.87M & 94.90\% & -- \\
    \textbf{SketchMamba (ours)} & \textbf{1.55M} & \textbf{94.93\%} & \textbf{0.706} \\
    \bottomrule
  \end{tabular}
\end{table}

\subsection{Ablation: The Role of Dense Supervision}
\label{ssec:res-ablation}

To isolate the driver of early prediction, we ablate the dense per-step
classification loss against a standard final-step-only loss. Training
curves (Fig.~\ref{fig:loss}) show both regimes converge to indistinguishable
final-step accuracy (Table~\ref{tab:results-ablation}). However, the
final-step-only model fails completely at progressive recognition, collapsing
the AUC from 0.706 to 0.074. The causal architecture alone is insufficient
for early prediction; dense supervision is strictly required to structure
the intermediate hidden states.

\begin{table}[t]
  \centering
  \caption{Ablation of the supervision regime using the SketchMamba backbone.}
  \label{tab:results-ablation}
  \small
  \setlength{\tabcolsep}{4pt}
  \begin{tabular}{@{}lcc@{}}
    \toprule
    \textbf{Supervision Regime} & \textbf{Final Top-1 Acc.} & \textbf{Prog. AUC} \\
    \midrule
    Final-step only & 94.89\% & 0.074 \\
    \textbf{Dense per-step (ours)} & \textbf{94.93\%} & \textbf{0.706} \\
    \bottomrule
  \end{tabular}
\end{table}

\subsection{Generative Fidelity and Latency}
\label{ssec:res-generation}

\begin{figure}[t]
  \centering
  \begin{minipage}[t]{0.48\textwidth}
    \centering
    \includegraphics[width=\linewidth]{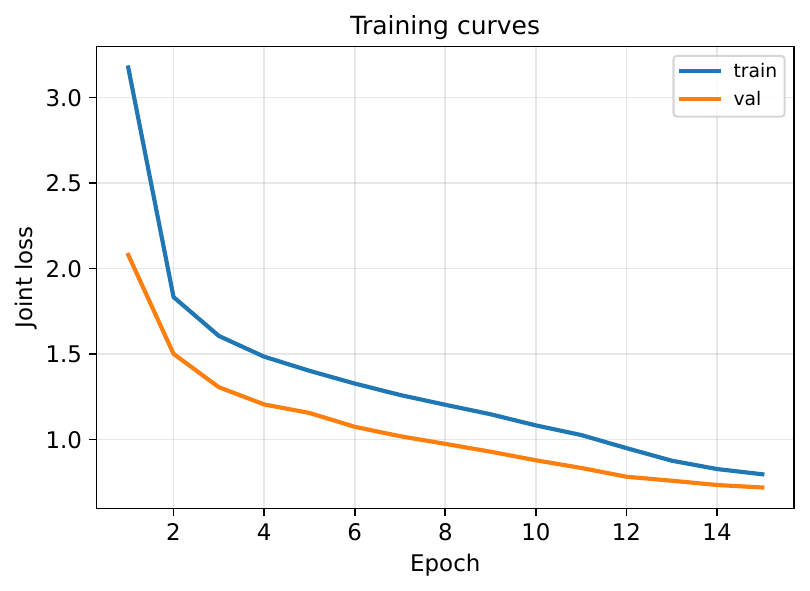}
    \caption{Training loss curves comparing dense vs.\ final-step supervision.}
    \label{fig:loss}
  \end{minipage}\hfill
  \begin{minipage}[t]{0.48\textwidth}
    \centering
    \includegraphics[width=\linewidth]{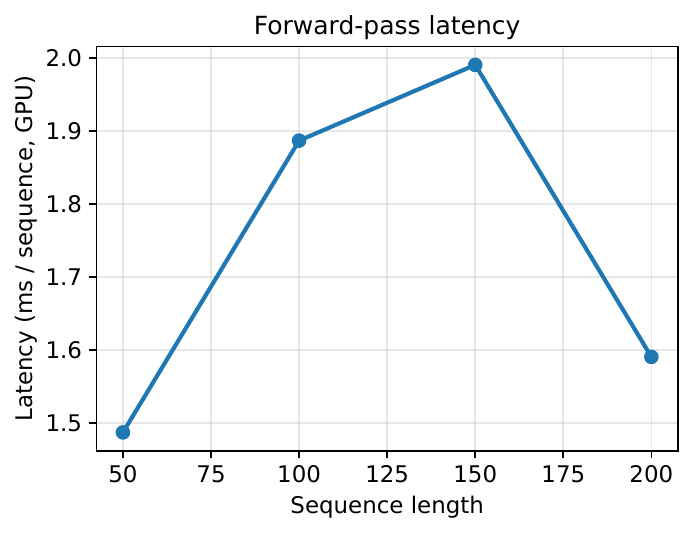}
    \caption{Inference latency scales in $O(1)$ with prefix length $L$.}
    \label{fig:latency}
  \end{minipage}

  \vspace{0.6em}
  \includegraphics[width=0.72\linewidth]{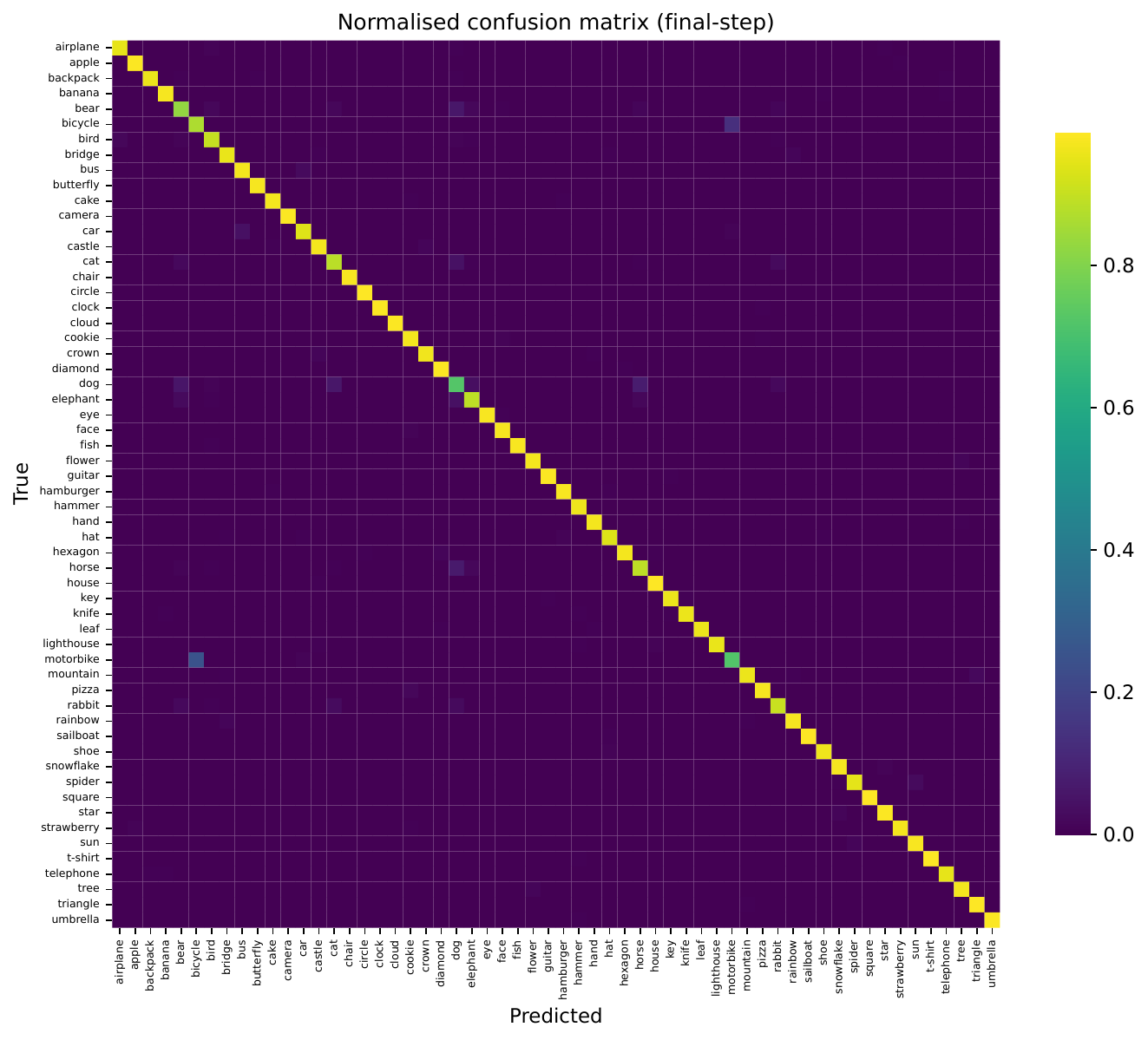}
  \caption{Confusion matrix detailing early-stroke ambiguities.}
  \label{fig:confusion}
\end{figure}

SketchMamba successfully conditions the bivariate GMM on the same causal
state used for recognition, generating semantically appropriate strokes
from arbitrary partial prefixes. Because the SSM updates its state in $O(1)$ time,
generation latency remains constant irrespective of prefix length
(Fig.~\ref{fig:latency}). On an NVIDIA L40S GPU, SketchMamba requires 17 MB
of peak memory and processes 1.5 ms per sequence, bypassing the $O(L)$
scaling overhead of the Transformer.

\subsection{Discussion and Limitations}
\label{ssec:res-discussion}

Analysis of the confusion matrix (Fig.~\ref{fig:confusion}) reveals
structured failure modes. Misclassifications cluster among visually similar
classes (e.g., circles versus clocks, or lines versus swords) where early
geometric primitives are identical. 

Furthermore, the standard cross-entropy loss struggles with pen-state
imbalance. Because pen-lift ($p_2$) and end-of-sketch ($p_3$) actions
account for a small fraction of total sequence steps compared to drawing
actions ($p_1$), the generative head occasionally fails to terminate
sketches cleanly. Replacing the uniform pen-state cross-entropy with a focal
loss to explicitly up-weight minority transition events represents a direct
path for future work.

%% file: sections/06_conclusion.tex
\section{Conclusion}
\label{sec:conclusion}

We introduced SketchMamba, a single causal sequence model that unifies
vector-sketch recognition and autoregressive generation. By replacing the
standard final-step classification loss with a dense per-step supervision
regime, we forced a causal selective state-space backbone to operate as an
early-prediction recogniser. The empirical results confirm that this
training regime, rather than specialised architectural branching, is the
active ingredient for progressive recognition. At 1.55 million parameters,
the model matches the final-step accuracy of a causal Transformer while
significantly outperforming recurrent and convolutional baselines, and it
processes sequence steps in constant time. 

Future work will extend this framework to larger vocabularies and address
the residual pen-state imbalance using transition-weighted objectives.
SketchMamba demonstrates that a single causal hidden state can fully
describe a drawing while it is actively being made, directly supporting
the requirements of streaming digital ink interfaces.

%% file: sections/07_credits.tex
\begin{credits}

\subsubsection{\discintname}
The authors have no competing interests to declare that are relevant to the
content of this article.
\end{credits}